\documentclass[lettersize,journal]{IEEEtran}
\usepackage{graphicx}
\usepackage{amsthm}
\usepackage{amsmath}
\usepackage{amsfonts,amssymb}
\usepackage{multirow}

\newtheorem{definition}{Definition}

\usepackage{bbding}
\usepackage{subfigure}
\usepackage{multirow}
\usepackage{booktabs}
\usepackage{makecell}
\usepackage{color}
\usepackage{stfloats}

\usepackage{array} \usepackage[caption=false,font=normalsize,labelfont=sf,textfont=sf]{subfig}
\usepackage{textcomp}
\usepackage{stfloats}
\usepackage{url}
\usepackage{verbatim}
\usepackage{longtable}
\usepackage{wrapfig}
\usepackage{cite}
\begin{document}
	
	\title{TodyNet: Temporal Dynamic Graph Neural Network for Multivariate Time Series Classification}
	
	\author{Huaiyuan Liu, Xianzhang Liu, Donghua Yang, Zhiyu Liang, Hongzhi Wang,~\IEEEmembership{Senior member, IEEE}, Yong Cui, and Jun Gu
		
		\thanks{Manuscript received April 11, 2023; revised August 16, 2021.}
		\thanks{This paper was supported by the Science and Technology Project of State Grid: Research on artificial intelligence analysis technology of available transmission capacity (ATC) of the key section under multiple power grid operation modes (5100-202255020A-1-1-ZN). (\emph{Corresponding author: Hongzhi Wang}.)}
		\thanks{Huaiyuan Liu, Donghua Yang, Zhiyu Liang, and Hongzhi Wang are with the Faculty of Computing, Harbin Institute of Technology, Harbin 150000, China (e-mail: 
			wangzh@hit.edu.cn).}
		\thanks{Xianzhang Liu is with the School of Energy Science and Engineering, Harbin Institute of Technology, Harbin 150000, China
			.}
		\thanks{Yong Cui and Jun Gu are with the State Grid Shanghai Municipal Electric Power Company, Shanghai 200000, China
			.}
	}
	
	\markboth{IEEE TRANSACTIONS ON NEURAL NETWORKS AND LEARNING SYSTEMS,~Vol.~xx, No.~xx, August~2023}%
	{Shell \MakeLowercase{\textit{et al.}}: A Sample Article Using IEEEtran.cls for IEEE Journals}
	
	\IEEEpubid{0000--0000/00\$00.00~\copyright~2021 IEEE}
	
	\maketitle
	
	\begin{abstract}
		Multivariate time series classification (MTSC) is an important data mining task, which can be effectively solved by popular deep learning technology. Unfortunately, the existing deep learning-based methods neglect the hidden dependencies in different dimensions and also rarely consider the unique dynamic features of time series, which lack sufficient feature extraction capability to obtain satisfactory classification accuracy. To address this problem, we propose a novel temporal dynamic graph neural network (TodyNet) that can extract hidden spatio-temporal dependencies without undefined graph structure. It enables information flow among isolated but implicit interdependent variables and captures the associations between different time slots by dynamic graph mechanism, which further improves the classification performance of the model. Meanwhile, the hierarchical representations of graphs cannot be learned due to the limitation of GNNs. Thus, we also design a temporal graph pooling layer to obtain a global graph-level representation for graph learning with learnable temporal parameters. The dynamic graph, graph information propagation, and temporal convolution are jointly learned in an end-to-end framework. The experiments on 26 UEA benchmark datasets illustrate that the proposed TodyNet outperforms existing deep learning-based methods in the MTSC tasks.
	\end{abstract}
	
	
	\section{Introduction}
	\IEEEPARstart{M}{ultivariate} Time Series (MTS) is ubiquitous in a wide variety of fields as a significant type of data, ranging from action recognition~\cite{li2021shapenet}, health care~\cite{kang2014bayesian} to traffic~\cite{ji2022stden}, energy management~\cite{wu2020connecting}, and other real-world scenario~\cite{song2021learning,yue2022ts2vec,sun2019hierarchical}. In the last decade, time series data mining has gradually become an important research topic, with the development of data acquisition and storage technology. The mining of time series data mainly covers three tasks: classification, forecasting, and anomaly detection. Multivariate Time Series Classification (MTSC) is the problem of assigning a discrete label to a multivariate time series, which assists people to judge the situation of the happened events.
	
	Multivariate time series classification is challenging. On the one hand, the characterization of MTS is more complex due to its various temporal order correlations and high dimensionality, compared to the tasks of univariate time series classification. On the other hand, unlike temporally invariant classification tasks, the hidden information contained in MTS is more abundant and more difficult to be mined.
	
	Numerous approaches have been proposed to settle the issue of MTSC over the years. Traditional methods have focused on distance similarity or features, such as Dynamic Time Warping with k-Nearest Neighbor (DTW-kNN)~\cite{seto2015multivariate} and Shapelets~\cite{ye2011time}, which have been validated to be effective on many benchmark MTS datasets. Nonetheless, the above-mentioned methodologies have to make the effort to data pre-processing and feature engineering due to the extraordinary difficulty of feature selection from the huge feature space. They fail to adequately capture the temporal dynamic relationships within the time series of each variable.
	
	\IEEEpubidadjcol 
	
	Recent research has turned to deep learning when solving end-to-end MTSC tasks because of the widespread applications of deep Convolutional Neural Networks (CNN)~\cite{c:2}, whose benefit is that deep learning-based method can learn low-dimensional features efficiently rather than dealing with huge amounts of feature candidates. Many methods utilized for MTSC adapted to the multivariate case by converting the models originally designed for univariate time series. Fully Convolutional Network (FCN)~\cite{wang2017time} has a better overall ranking compared with traditional approaches. MLSTM-FCN~\cite{karim2019multivariate} augments the FCN with a squeeze-and-excitation block to achieve better performance of classification. In addition, some approaches are specifically dedicated to MTSC by learning the latent features. These types of deep learning-based strategies have shown gratifying results in MTSC tasks. Unfortunately, current deep-learning-based models for the issue of MTSC rarely consider the hidden dependency relationships between different variables.

	The dependency relationships can be naturally modeled as graphs. In recent years, Graph Neural Networks (GNNs) have been successfully utilized to processing of graph data by their powerful learning capability for graph structure. The neighborhood information of each node can be captured through the diverse structural information propagation of the graph neural network. Thrillingly, each variable can be characterized as a node in a graph for MTS, which are interconnected through hidden dependencies. Therefore, it is a promising way that models MTS data by graph neural networks to mine the implicit dependencies between variables on the time trajectory. Spatial-temporal graph neural networks take an external graph structure and time series data as inputs, which can significantly improve the performance of the methods that do not exploit graph structural information~\cite{han2021dynamic}. However, only one graph is constructed over the entire temporal trajectory, which limits and neglects the impact of dynamic processes on representational ability. Thus, the following key issues and challenges should be valued.
	
	\begin{itemize}
		\item\textit{Mining Hidden Dependencies.} Most models for MTSC focus on the extraction of inter-variable features but neglect the possibility that implicit dependencies between different variables. The question then is how to design an effective framework to capture these information.
		\item\textit{Dynamic Graph Learning.} Most existing GNNs approaches rely heavily on predefined graph structures and only emphasize message propagation without considering the dynamic properties of time series data. Thus, how to dynamically learn the spatial-temporal features and graph structure of MTS is also an important issue.
		\item\textit{Temporal Graph Pooling.} Most deep learning methods for MTSC pool a large number of dimensions directly at the end of models, which is called flat and may cause missing information. Some graph pooling methods have used GNNs to alleviate it~\cite{zhang2018end, ying2018hierarchical, lee2019self}, however, they did not consider the internal temporal features.
	\end{itemize}
	
	To address the above issues, we propose a novel \textbf{T}emp\textbf{o}ral \textbf{Dy}namic Graph Neural \textbf{Net}work (TodyNet) for multivariate time series classification. For \emph{mining hidden dependencies}, we propose a novel end-to-end framework that discovers the dependence correlations between variables, characteristics within variables, and spatial-temporal dependencies of variables by graph construction and learning, temporal convolution, and dynamic graph neural network, respectively. For \emph{dynamic graph learning}, the internal graph structure can be learned by gradient descent, and we design a graph transform mechanism to propagate the dynamic properties of multivariate time series. For \emph{temporal graph pooling}, a hierarchical temporal pooling approach is proposed to avoid the flat and achieve high performance. In summary, the main contributions of our work are as follows:
	
	\begin{itemize}
		\item \textit{\textbf{Pioneering.}} To the best of our knowledge, this is the first study for multivariate time series classification based on the temporal dynamic graph.
		\item \textit{\textbf{Novel Joint Framework.}} An end-to-end framework for MTSC is proposed to jointly learn the dynamic graph, graph propagation, and temporal convolution.
		\item \textit{\textbf{Dynamic Graph Mechanism.}} A dynamic graph processing mechanism is proposed to capture the hidden dynamic dependencies among different variables and between adjacent time slots and enhance the effectiveness of graph learning.
		\item \textit{\textbf{Temporal Graph Pooling.}} A brand-new temporal graph pooling is presented to avoid the flat of pooling methods, and combined with temporal features extraction simultaneously.
		
		\item \textit{\textbf{Remarkable Effect.}} Experimental results show that our method significantly improves the performance of mainstream deep learning classifiers and outperforms state-of-the-art methods.
	\end{itemize}
	
	The remainder of this paper is organized as follows. Section~\ref{s2} briefly introduces the related works. Section~\ref{s3} is devoted to presenting the problem formulation, the notation, and the definition of the related concept. The details of the proposed TodyNet are described in Section~\ref{s4}. The experimental results and ablation studies are illustrated in Section~\ref{s5}, which is followed by the conclusion.
	
	\section{Related Work}
	\label{s2}
	In this section, we briefly summarize the recent advances in
	multivariate time series classification tasks and spatial-temporal graph neural networks.
	
	\subsection{Multivariate Time Series Classification}
	The multivariate time series classification problem expects to utilize time series data with multivariate features to accurately predict a number of specific classes~\cite{bianchi2020reservoir,lin2017gcrnn}. In general, distance-based, feature-based, and deep-learning-based methods are the three major types of approaches to address MTSC tasks.  Multichannel Deep Convolutional Neural Network (MCDCNN)~\cite{zheng2014time} captures features within variables by 1D convolution and then combined them with a fully connected layer, which is a pioneering approach that applied CNN to MTSC. Inspired by this, many effective network architectures began to spurt erupted. Time Series Attentional Prototype Network (TapNet)~\cite{c:3} learns the low-dimensional features by a random group permutation method and constructs an attentional prototype network to overcome the issue of limited training labels. Currently, OS-CNN~\cite{c:2} is the latest model that can achieve state-of-the-art performance by designing an Omni-Scale block to cover the best receptive field size across different datasets. Nevertheless, deep-learning-based models have obvious limitations. They assume that the dependencies between different variables are the same, resulting in the pairwise relationships between variables cannot be effectively represented. In this case, the graph is the most appropriate data structure to model the MTS.
	
	\subsection{Graph Neural Networks}
	Since the concept of Graph Neural Networks (GNNs) was introduced~\cite{scarselli2008graph}, there has been an explosive development in the processing of graph information. GNNs follow a local aggregation mechanism in that the embedding vector of each node can be computed by recursively aggregating and transforming the information of its neighbors. In recent years, various variants of GNNs have been proposed~\cite{wu2020comprehensive,chen2018fastgcn,xu2018powerful,morris2019weisfeiler}. For example, the graph convolution network (GCN)~\cite{kipf2016semi} is a representative work that extends convolution to the spectral domain by finding the corresponding Fourier basis. GraphSAGE~\cite{hamilton2017inductive} and the Graph Attention Network (GAT)~\cite{velickovic2017graph} are also typical approaches that generalize typical convolution to spatial neighbors. In order to simultaneously meet the need for spatial and temporal dimensions of the data, a new family of GNNs has been born, i.e. spatial-temporal GNNs, which can learn the graph structure over temporal trajectories~\cite{cao2020spectral,chen2022tamp,guo2021hierarchical}. For example, ~\cite{li2021spatial} proposed a new model named STFGN for fusing newly generated graphs and given spatial graphs to capture hidden spatial dependencies and learn spatial-temporal features. Regretfully, it is difficult to find a predefined graph structure for the MTSC tasks. In addition, most of the current spatial-temporal GNNs are designed for traffic forecasting, there is a gap in the MTSC tasks that need to be filled.
	
	\section{Preliminaries}
	\label{s3}
	
	\begin{figure*}[!t]
		\centering
		\includegraphics[width=1\textwidth]{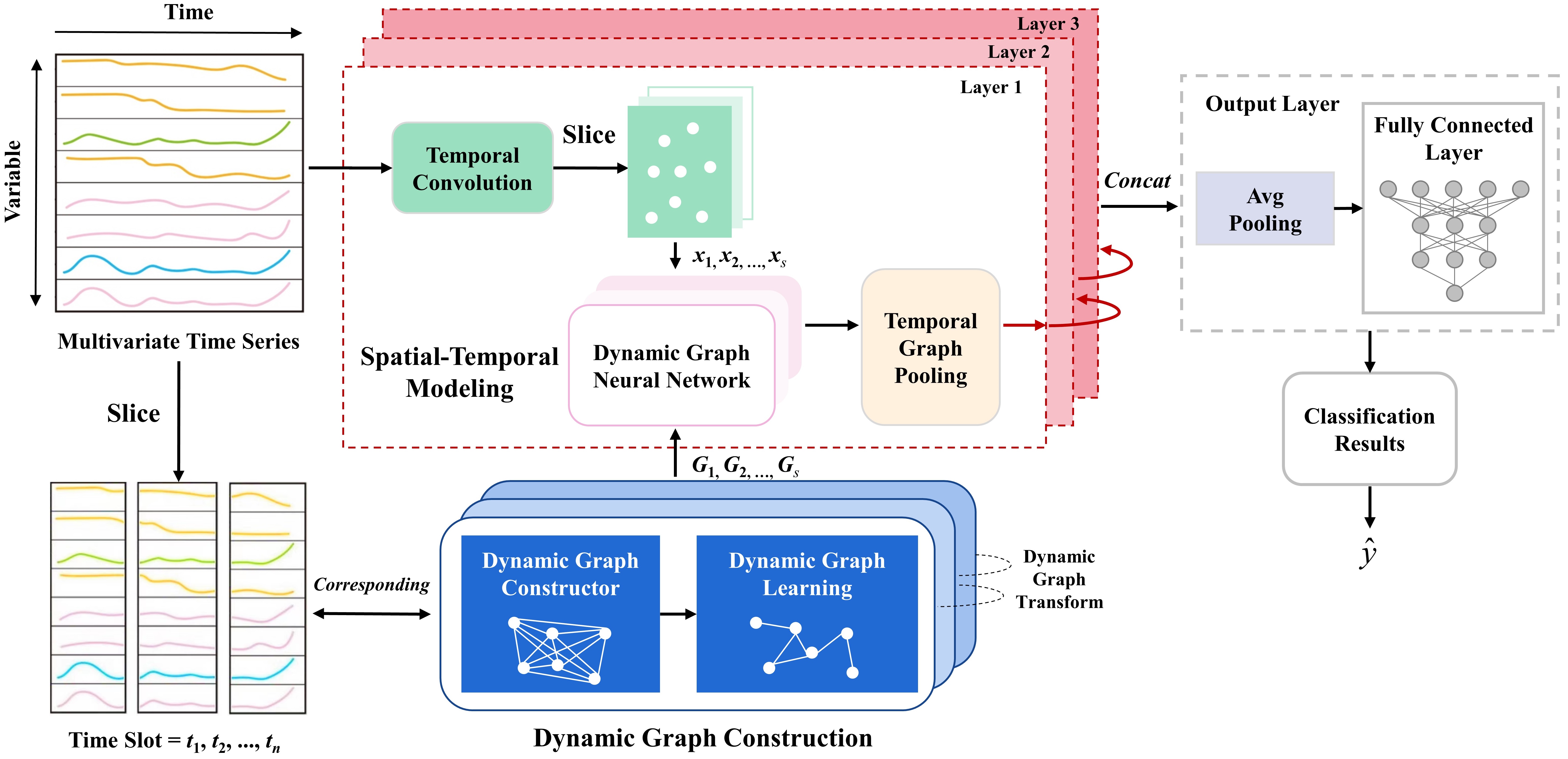}
		\caption{The framework of TodyNet. We first split the input time series into $s$ slices, and generate a dynamic graph for each slice. The dynamic graph neural network modules and temporal convolution modules capture spatial and temporal dependencies separately. Afterward, the temporal graph pooling module clusters nodes together with learnable temporal parameters at each layer. The output layer processes concatenate hidden features for final classification results.}
		\label{fig:framework}
	\end{figure*}
	
	This section defines the key concept and notations of this paper. To begin with, we formulate the problem that we are interested in, multivariate time series classification.
	
	\subsection{Problem Formulation}
	A Multivariate Time Series (MTS) $X = \{ x_1, x_2,...,x_d\}\in \mathbb{R}^{d\times l}$ denote the value of multivariate variable of dimension $d\in \mathbb{N^*}$ at the time series with $x_i=\{x_{i1}, x_{i2},..., x_{il}\}$, $i=1, 2,.., d$, where $l \in \mathbb{N^*}$ denote the length of the multivariate time series. Given a group of multivariate time series $\chi = \{ X_1, X_2,..., X_m\}\in \mathbb{R}^{m\times d\times l}$ and the corresponding labels $\eta=\{ y_1, y_2,...,y_m\}$, where $y$ is a predefined class label of each multivariate time series, and $m\in \mathbb{N^*}$ is the number of time series. The MTSC tasks are dedicated to predicting unlabeled multivariate time series by training a classifier $f(\cdot)$ from $\chi$ to $\eta$.
	
	\subsection{Graph Notations and Definitions}
	\noindent\textbf{Graph-related Concepts.} Graph generally describes relationships among entities in a network. In the task of multivariate time series classification, we give formal definitions of temporal graph-related concepts below.
	
	\begin{definition}[Temporal Graph]
		The temporal graph is constructed by multivariate time series. A temporal graph is given in the form $G=(V, E)$ where $V=\{v_1,...,v_n\}$ is the set of nodes with a number of $n$, and $E$ is the set of edges. The nodes denote the variables, and the edges denote the relationship among different variables defined as similarity or some structure learning results.
	\end{definition}
	
	
	\noindent\textbf{Static and Dynamic Graph.} The static and dynamic graphs are defined by the temporal dependency of graph construction. Each multivariate time series $X$ is segmented averagely by some isometric time slots $T = \{ T_1, T_2,..., T_S\}$ in chronological order, where $S \in \mathbb{N}^*$ is the number of time slot. Generally, each time slot is denoted by $T_i = \{t_{1+(i-1)s}, ..., t_{is}\}$, $i = 1, ..., s$, $\left| T_i \right| = s > 0$. Given a set of temporal graphs $G_T = \{V, E_T\}$ with the same nodes but different edges at each time slot, where $G_T = \{G_{T_1}, G_{T_2},  ..., G_{T_S}\}$, $E_T = \{E_{T_1}, E_{T_2},  ..., E_{T_S}\}$.
	
	\begin{definition}[Static Graph]
		The static graph is a temporal graph for each multivariate time series with $S=1$, $i.e.$ there is no segmentation existence for each multivariate time series, and $s$ is equal to the length of the whole time series. The size of sets $\left| G_T \right| = \left| E_T \right| = 1$.
	\end{definition}
	
	\begin{definition}[Dynamic Graph]
		\label{def:dygraph}
		The dynamic graph is a set of temporal graphs for each multivariate time series with $S > 1$, $i.e.$ edges of the temporal graph have dynamic variations at different time slots. The size of sets $\left| G_T \right| = \left| E_T \right| > 1$.
	\end{definition}
	
	The dynamic variations of edges mean that the weight of edges and the connection relationship of vertices changes on different time slots while vertices stay statically constant for each multivariate time series.
	
	\section{Proposed Model}
	\label{s4}
	
	In this section, we will introduce our new proposed TodyNet model.
	We first introduce the overall framework of TodyNet in Section \ref{subsec:overall}. Then we demonstrate each component of our model in Section \ref{subsec:gc&l} through Section \ref{subsec:tc}.
	
	\subsection{Model Architecture Overview}
	\label{subsec:overall}
	We first introduce the general framework of our proposed deep learning-based model TodyNet briefly. As illustrated in Figure~\ref{fig:framework}, TodyNet is composed of the following core modules. To establish the initial relationships between different dimensions, a \emph{graph construction and learning module} generates a set of graph adjacency matrices, which correspond to different time slots respectively. And the elements of graph adjacency matrices can be learned with the iteration process. $k$ \emph{Dynamic graph neural network modules} consists of dynamic graph transform and dynamic graph isomorphism network. Dynamic graph transform exploits the dynamic associations between different temporal graphs, whose results are implied in the new adjacency matrices. Dynamic graph isomorphism network processes time series data with $k$ \emph{temporal convolution modules}, which captures spatial-temporal dependencies of multivariate time series.
	
	To avoid the flat pooling problem, the \emph{temporal graph pooling module} designed a differentiable and hierarchical graph pooling approach with learnable temporal convolution parameters. To obtain the final classification result, the \emph{output module} uses average pooling and a fully connected layer to get the values for each category. Our model will be introduced in detail in the following subsections.
	
	\subsection{Construction and Learning of Temporal Dynamic Graph}
	\label{subsec:gc&l}
	Since there is no predefined graph structure for universal time series data, we will start by demonstrating a general graph-constructed module that generated adjacency matrices to create the initial associations of dimensions. Besides, the hidden dependencies between variables are represented by graph adjacency matrices and optimized with training iterations. For the sake of simplicity, we do not establish this relationship through time series data and encode the adjacency matrices by ``shallow" embedding. Each node is assigned two values, which represent the source and target nodes, respectively. Therefore, we generated two vectors $\Theta$ and $\Psi$ with length $d$ for each time slot $t$, and all elements are learnable parameters that are initialized randomly.
	Then we computed the multiplication of $\Theta^{{\rm T}}$ and $\Psi$ whose result will be regarded as the initial adjacency matrix for a time slot. And the values of the adjacency matrix can be optimized by training. The graph construction is illustrated as follows:
	
	\begin{align}
		A & = \Theta^{\rm T} \cdot \Psi \\
		\label{eq:topk}
		idx,idy & = argtopk(A[:,:]) \quad idx \neq idy\\
		\label{eq:spares}
		A[& -idx, -idy] = 0
	\end{align}
	
	where, $\Theta = [\theta_{t,1}, \theta_{t,2}, ..., \theta_{t,d}]$, $\Psi = [\psi_{t,1}, \psi_{t,2}, ..., \psi_{t,d}]$ represent the random initialization of learnable node embeddings, and $argtopk(\cdot) $ returns the indices of the top-k largest values of the adjacency matrix $A$. The
	adjacency matrix is sparsized by Equation \ref{eq:topk}-\ref{eq:spares}, which enables reducing the computation cost. For the adjacency matrix of each time slot, we only preserve the elements with the top-k largest weights and set other values as zero.
	
	\subsection{Dynamic Graph Neural Network}
	\label{subsec:gnn}
	We purpose to explore the spatial relationships between different time series dimensions and represent the interaction of their features in graphs. The dynamic graph neural network module focuses on message passing and feature aggregating between nodes. Though some GNNs models are able to propagate neighbor messages, the existing methods can only operate on static graphs. In this paper, we propose a novel model that processes message propagation for dynamic graphs based on an improved Graph Isomorphism Network (GIN).
	
\begin{figure}[!ht]
		\centering
		\includegraphics[width=1\columnwidth]{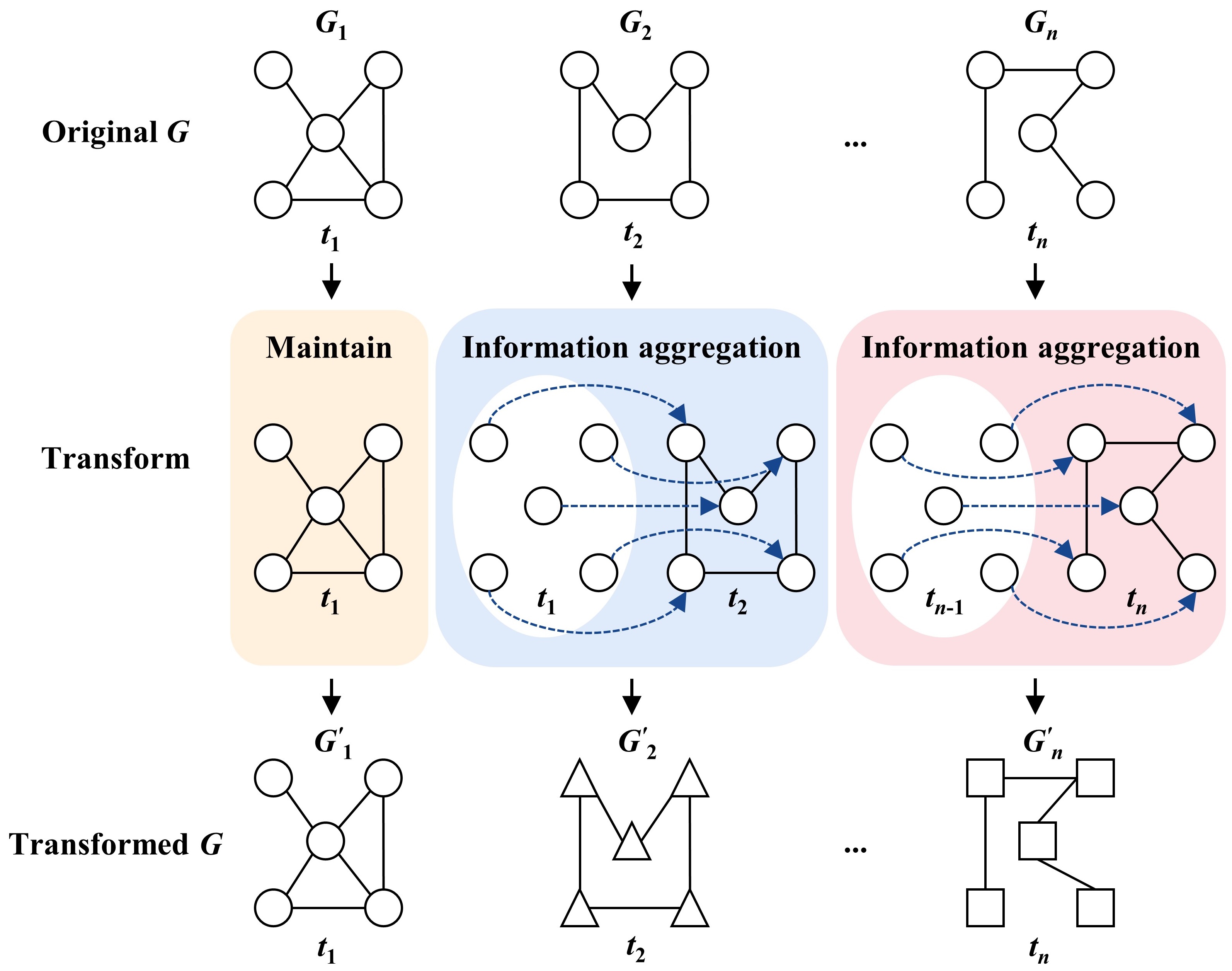}
		\caption{Dynamic Graph Transform. The latter graph aggregates information from the previous graph for corresponding nodes.}
		\label{fig:trans}
	\end{figure}

	\noindent\textbf{Dynamic Graph Transform.}
	We design a transformation on the set of dynamic graphs, which establishes the associations between different graphs. The basic assumption behind these time slots is that the later time slots data is transformed from the data in the earlier time slots. To discretization, we attempt to construct a connection with the graph corresponding to the previous time slot for each graph. The structure of Dynamic Graph Transform(DGT) is illustrated in Figure~\ref{fig:trans}.
	
	For each graph of time slot, the same number of vertices are added which characterize the data of the previous time slot graph of the corresponding vertices respectively, except for the first graph. Therefore, the set of vertices will be modified to $\{ v_{(t,1)},\, v_{(t,2)}, ...,\,  v_{(t,N)},\,\, v_{(t-1,1)},\, v_{(t-1,2)}, ...,\, v_{(t-1,N)} \}$.
	We assign orientations ($i.e.$ \emph{directed edges}) from the previous time vertices to their counterparts at the current time, which signifies adding connections from $v_{(t-1,n)}$ to $v_{(t,n)}$ for $n=1,2,..., N$. Then the latest generated set of graphs is able to be transported to the next steps.
	
	Actually, it is unnecessary to double the number of vertices in practical implementation. We can aggregate source nodes embedding to target nodes embedding directly for new directed edges, then delete the new vertices.
	
	\noindent\textbf{Dynamic Graph Isomorphism Network (DyGIN).}
	\label{subsubsec:gin}
	Graph Isomorphism Network(GIN) is one of the powerful GNNs, which can discriminate different graph structures that other popular GNNs variants cannot distinguish. Motivated by GIN, we designed a novel method to aggregate information that consists of a dynamic paradigm through parallelism as Equation~\ref{eq:parallel}. In contrast to static GNNs, dynamic GNNs separate different time slot data in the same dimension completely after DGT. Actually, the same vertex at different time slots will aggregate information from a different set of vertices in general. The DyGIN can be defined as:
	
	\begin{align}
		h_v^{(l,\, t)} = & MLP^{(l,\, t)} \Big( \big( 1 + \epsilon^{(l)} \big) \cdot h_v^{(l-1,\, t)} + h_v^{(l-1,\, t-1)} + \notag \\
		& \sum_{u \in \mathcal{N}(v)} \tilde{\omega}_{ij} \cdot h_u^{(l-1,\, t)} \Big)
	\end{align}
	
	\begin{equation}
		\label{eq:parallel}
		h_v^{(l)} = CONCAT \Big( h_v^{(l,\, t)} \enspace \big\lvert \enspace t = 0, 1, ..., T \Big)
	\end{equation}
	
	where, $h_v^{(l,\, t)}$ represents the output of GIN for node $v$ at $t$ time slot in $l$ layer, $h_v^{(l-1,\, t-1)}$ is the simple implementation for DGT, which is available for $t > 2$ (the second and subsequent graphs). The edge weight, $\omega_{ij}$, is normalized to $\bar{\omega}_{ij}$. We can regard $\epsilon$ as a learnable parameter.
	
	We also provide the adjacency matrix form:
	
	\begin{equation}
		H^l = MLP \Big( \big( \tilde{A} + (1 + \epsilon^{l}) \cdot I \big) \cdot H^{l-1} + H^{l-1}[t_1:t_{T-1}] \Big)
	\end{equation}
	
	where, $H^{l-1}$ represents the output tensor of $l$-th GIN layer, $H^{l-1}[t_1:t_{T-1}]$ is aligned from the second time slot data, $\tilde{A}$ is normalized through $D^{-\frac{1}{2}} A D^{\frac{1}{2}}$, and $D$ is the degree matrix of $A$.
	
	\subsection{Temporal Graph Pooling}
	\label{subsec:gp}
	In general, pooling is a necessary procedure after the classifiers have generated new features. However, some universal approaches of pooling, such as max pooling, mean pooling, and sum pooling, aggregate a vector even a matrix of features into a few features, which is \emph{flat} and may lose much information. To the best of our knowledge, there is no such method that deals with this issue in time series analysis. In this section, we propose a novel Temporal Graph Pooling (TGP) that combined graph pooling and temporal process, which can alleviate the difficulty.
	
\begin{figure}[ht]
		\centering
		\includegraphics[width=1\columnwidth]{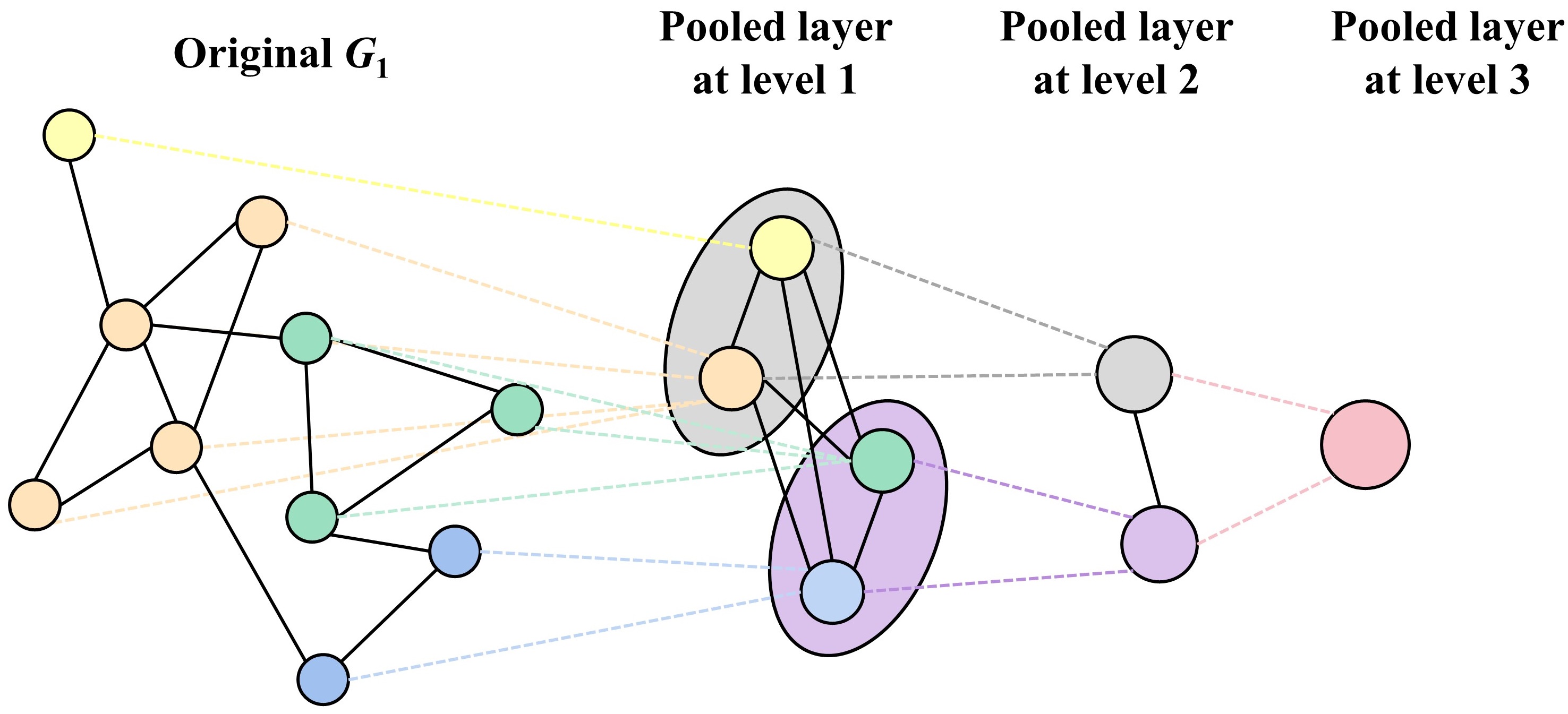}
		\caption{An abstract illustration of Temporal Graph Pooling. At each layer, we utilize time convolution to cluster nodes and extract the temporal features. Then we reconstruct adjacency matrices through convolution weights.}
		\label{fig:pool}
	\end{figure}

	
	Temporal Graph Pooling provided a solution through the \emph{hierarchical} pooling method. The core idea is to Learn to assign nodes to clusters, which can control the number of reducing nodes. We applied it hierarchically after each GNN layer, as illustrated in Figure~\ref{fig:pool}.
	
	\noindent\textbf{Pooling with convolution parameters.}
	A 2-dimensional convolution neural network(CNN) layer is designed to concentrate nodes into clusters according to the given parameter \emph{pooled ratio}, and nodes are considered as the channels that extract features in CNN. In order not to break the window of the receptive field, the kernel size of temporal convolution is assigned to this convolution kernel size. $X_l$ represents the input node embedding tensor at $l$-th layer, the CNN to compute the output embedding tensor $X_{l+1}$ can be denote as:
	
	\begin{equation}
		\label{eq:cnn}
		X^{l+1} = \sum_{j=0}^{N^l - 1} \text{weight}(N^{l+1}, j) \star X^l + \text{bias}(N^{l+1})
	\end{equation}
	
	where, $\star$ is the valid 2D cross-correlation operator, $N$ denotes the number of nodes in or out pooling module, note that $N^l$ equals to in\_nodes and $N^{l+1}$ equal to out\_nodes for $l$-th layer.
	
	After the generation of output tensor $X^{l+1}$, we consider the approach to compute the corresponding adjacency matrix. There is an observation that the shape of learnable weights $W^l$ is [$N^{l+1}$, $N^l$, 1, kernel\_size], then the vector $V^l \in \mathbb{R}^{1 \times k}$ composed of learnable parameters at layer $l$ is generated. A learnable assigned matrix $M^l = W^l \cdot V^l \in \mathbb{R}^{N^{l+1} \times N^l}$ can be obtained that the rows correspond to $N^{l+1}$ nodes or clusters and the columns corresponds to $N^l$ clusters. Given matrix $M$ and the adjacency matrix $A^{(l)}$ for input data at this layer, the following equation is utilized to generate output adjacency matrix $A^{(l+1)}$:
	
	\begin{equation}
		\label{eq:gp}
		A^{(l+1)} = M^{(l)} A^{(l)} M^{{(l)}^T} \in \mathbb{R}^{N_{l+1} \times N_{l+1}}
	\end{equation}
	
	Equation~\ref{eq:cnn}-\ref{eq:gp} provides the overall steps of Temporal Graph Pooling(TGP). In Equation~\ref{eq:cnn}, $X^{l+1}$ represents the output clusters embeddings after aggregating input embeddings. In Equation \ref{eq:gp}, $A^{(l+1)}$ denotes the connection relationships and corresponding weights of new clusters. In addition, each element $A^{(l+1)}_{ij}$ indicates the connected weight between $i$ and $j$. Therefore, TGP implements the hierarchical and differentiable graph pooling with temporal information, while optimizing the clusters aggregating approach during training.
	
	\subsection{Temporal Convolution}
	\label{subsec:tc}
	The temporal convolution (TC) module focuses on the capture of temporal dependencies within each dimension. In contrast to other deep learning methods, to highlight the advantage of graph learning modules, there is no complex processing in this module. The 3 convolution neural network layers with different convolution kernels are employed and neither \emph{dilation} nor \emph{residual connection} are used for simplicity. However, \emph{Padding} is applied to extend the length of the time series, which makes the output of time convolution equal in length to the input data. The output tensor of temporal features extracting will be transported to the GNN module after dividing according to the corresponding time slots.
	
	\begin{table*}[!b]
		\caption{Basic information of 26 UEA datasets.\label{tab:long}}
		\centering
		\begin{tabular}{c c c c c c c c c}
			\hline
			Code & Datasets & Type & Num Series & Series Length & Num Classes & Train Size & Test Size \\ \hline
			AWR & ArticularyWordRecognition & MOTION & 9 & 144 & 25 & 275 & 300 \\
			AF & AtrialFibrillation & ECG & 2 & 640 & 3 & 15 & 15 \\
			BM & BasicMotions & HAR & 6 & 100 & 4 & 40 & 40 \\
			CR & Cricket & HAR & 6 & 1197 & 12 & 108 & 72 \\
			DDG & DuckDuckGeese & AUDIO & 1345 & 270 & 5 & 50 & 50 \\
			EW & EigenWorms & MOTION & 6 & 17984 & 5 & 128 & 131 \\
			EP & Epilepsy & HAR & 3 & 206 & 4 & 137 & 138 \\
			EC & EthanolConcentration & OTHER & 3 & 1751 & 4 & 261 & 263 \\
			ER & ERing & HAR & 4 & 65 & 6 & 30 & 270 \\
			FD & FaceDetection & EEG & 144 & 62 & 2 & 5890 & 3524 \\
			FM & FingerMovements & EEG & 28 & 50 & 2 & 316 & 100 \\
			HMD & HandMovementDirection & EEG & 10 & 400 & 4 & 160 & 74 \\
			HW & Handwriting & HAR & 3 & 152 & 26 & 150 & 850 \\
			HB & Heartbeat & AUDIO & 61 & 405 & 2 & 204 & 205 \\
			LIB & Libras & HAR & 2 & 45 & 15 & 180 & 180 \\
			LSST & LSST & OTHER & 6 & 36 & 14 & 2459 & 2466 \\
			MI & MotorImagery & EEG & 64 & 3000 & 2 & 278 & 100 \\
			NATO & NATOPS & HAR & 24 & 51 & 6 & 180 & 180 \\
			PD & PenDigits & MOTION & 2 & 8 & 10 & 7494 & 3498 \\
			PEMS & PEMS-SF & MISC & 963 & 144 & 7 & 267 & 173 \\
			PS & PhonemeSpectra & SOUND & 11 & 217 & 39 & 3315 & 3353 \\
			RS & RacketSports & HAR & 6 & 30 & 4 & 151 & 152 \\
			SRS1 & SelfRegulationSCP1 & EEG & 6 & 896 & 2 & 268 & 293 \\
			SRS2 & SelfRegulationSCP2 & EEG & 7 & 1152 & 2 & 200 & 180 \\
			SWJ & StandWalkJump & ECG & 4 & 2500 & 3 & 12 & 15 \\
			UW & UWaveGestureLibrary & HAR & 3 & 315 & 8 & 120 & 320 \\ \hline
		\end{tabular}
	\end{table*}
	
	\section{Experimental Studies}
	\label{s5}
	In this section, extensive experiments have been conducted on the UEA benchmark datasets for multivariable time series classification to show the performance of TodyNet. Furthermore, we validate the performance of key components of TodyNet with a series of ablation experiments. Moreover, we visualize the class prototypes and time series embeddings to illustrate the excellent representation capabilities of TodyNet.
	
	\subsection{Experimental Settings}
	\noindent\textbf{Datasets.} UEA multivariate time series classification archive\footnote{https://www.timeseriesclassification.com} collected from different real-world applications covers various fields. We exclude the datasets with unequal length or missing values in the UEA multivariate time series classification archive~\cite{r:1}, to ensure the rationality of experiments. Thus, we use all 26 equal-length datasets of the 30 total and most implementations are similar setups in other studies. The details statistics of the benchmark datasets are as follows:
	The UEA multivariate time series classification archive is composed of real-world multivariate time series data, and we choose 26 data of equal length in the time series of each dimension as the datasets.  The length range of the datasets is from 8 to 17,984, and the range of dimension values is from 2 to 1345. The details of each dataset are shown in Table~\ref{tab:long}.
	
	\noindent\textbf{Baselines.} To validate the performance of TodyNet, the state-of-the-art or most popular approaches are selected as baselines for comparison. Baseline methods for multivariate time series classification are summarized as follows:
	
	(1) \textbf{OS-CNN} and \textbf{MOS-CNN}~\cite{c:2}: One-dimensional convolutional neural networks covered the receptive field of all scales, which are the latest and best deep learning-based methods achieved the highest accuracy for time series classification.
	
	(2) \textbf{ShapeNet}~\cite{li2021shapenet}: The latest shapelet classifier that embeds shapelet candidates with different lengths into a unified space.
	
	(3) \textbf{TapNet}~\cite{c:3}: A novel model that combines the benefits of traditional and deep learning, which designs a framework containing a LSTM layer, stacked CNN layer, and attentional prototype network.
	
	(4) \textbf{WEASEL+MUSE}~\cite{schafer2017multivariate}: The most effective bag-of-patterns algorithm that builds a multivariate feature vector.
	
	(5) \textbf{WLSTM-FCN}~\cite{karim2019multivariate}: A famous deep-learning framework obtained the representations by augmenting LSTM-FCN with
	squeeze-and-excitation block.
	
	(6) \textbf{ED-1NN}~\cite{c:2}: One of
	the most popular baselines based on Euclidean Distance and the nearest neighbor classifier. 
	
	(7) \textbf{DTW-1NN-I}~\cite{c:2}: One of
	the most commonly used baselines that process each dimension independently by dynamic time warping with the nearest neighbor classifier.
	
	(8) \textbf{DTW-1NN-D}~\cite{c:2}: Another similar baselines that process all dimensions simultaneously.
	
	\noindent\textbf{Implementation details.} In our model, the number of dynamic graphs for MTS is 4, the $k$ value for top-$k$ is set to 3 and the pooled ratio is 0.2. There are 3 layers of temporal convolution, and the convolution kernel size is 11,3,3. The channel size of temporal convolution is 64,128,256. The batch size is set to 16, and the learning rate is set to $10^{-4}$. We tune the kernel size for a few datasets due to the large differences in the dimension and length of each dataset. All the experiments are implemented on Pytorch 1.11.0 in Python 3.9.12 and trained with 2,000 epochs (computing infrastructure: Ubuntu 18.04 operating system, GPU NVIDIA GA102GL RTX A6000 with 48 Gb GRAM)\footnote{The source code is available at https://github.com/liuxz1011/TodyNet}.
	
	\noindent\textbf{Evaluation metrics.} We evaluate the performance of multiple classifiers over multiple test datasets by computing the accuracy, average accuracy, and the number of Wins/Draws/Losses. In addition, we also construct an adaptation of the critical difference diagrams~\cite{r:14}, replacing the posthoc Nemenyi test with the pairwise Wilcoxon signed-rank tests, and cliques formed using the Holm correction recommended~\cite{r:15,r:16}.
	
	\begin{figure*}[t]
		\centering    
		{\begin{minipage}{0.32\textwidth}
				\centering    
				\includegraphics[scale=0.42]{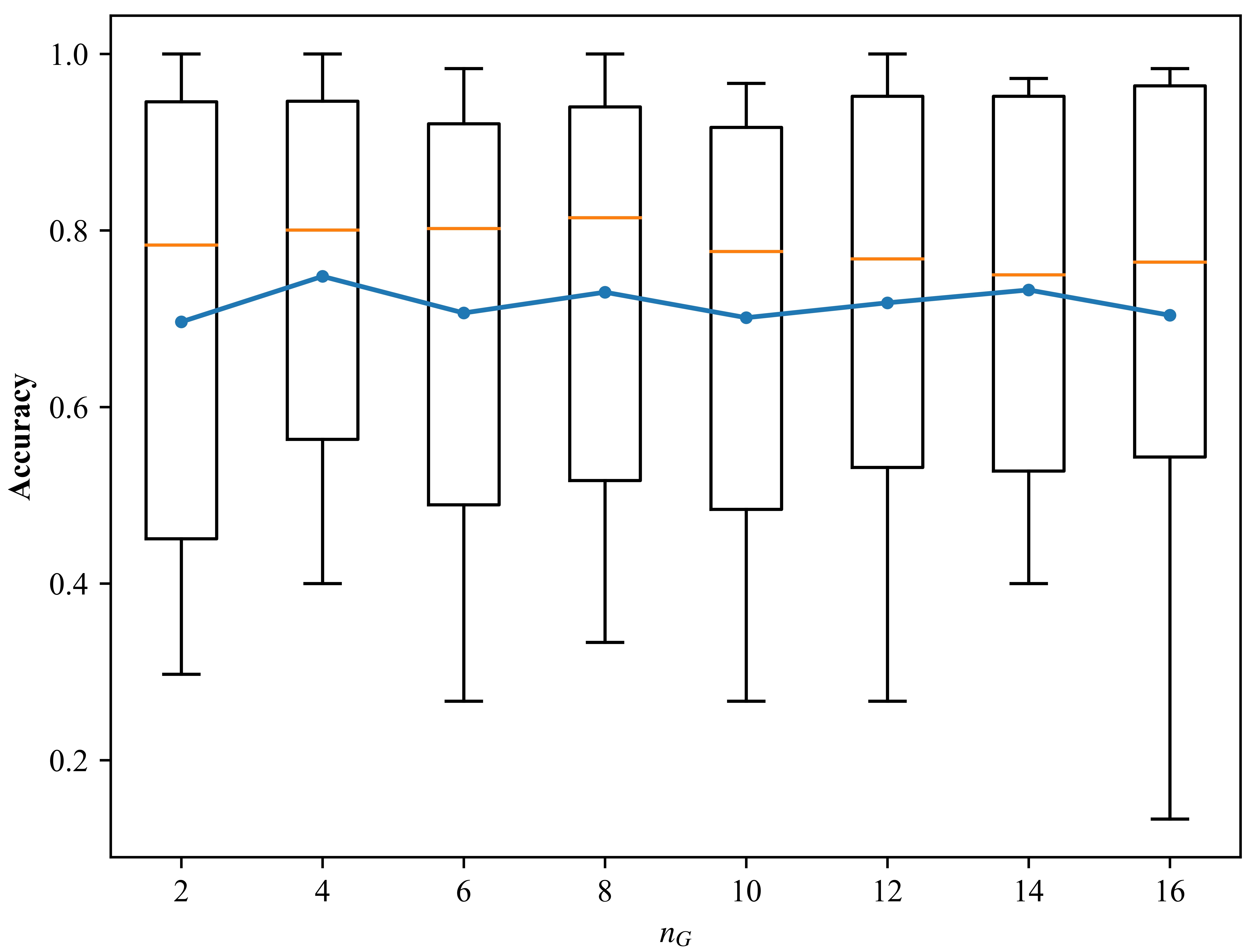}   
		\end{minipage}}	
		{\begin{minipage}{0.32\textwidth}
				\centering    
				\includegraphics[scale=0.42]{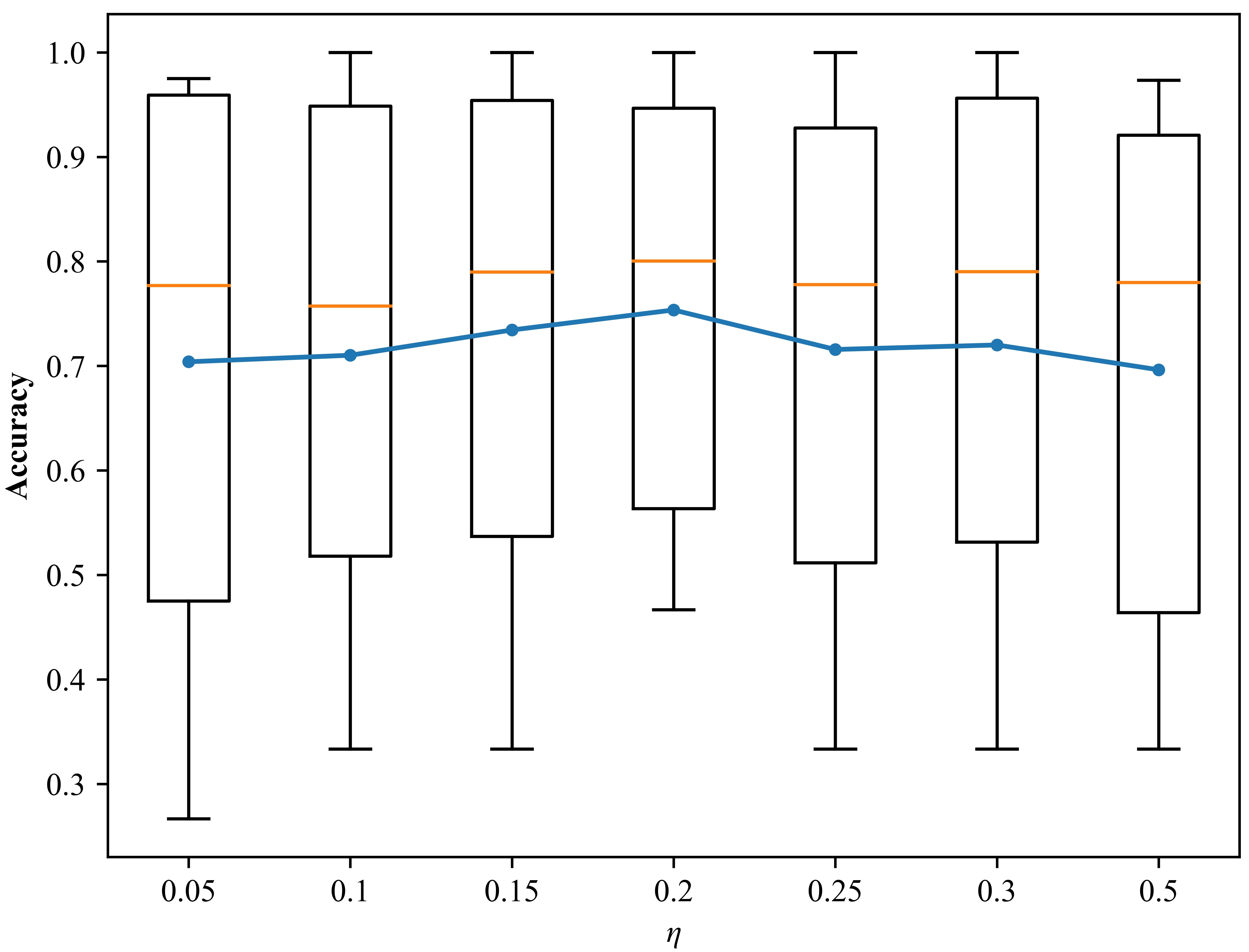}   
		\end{minipage}}
		{\begin{minipage}{0.32\textwidth}
				\centering    
				\includegraphics[scale=0.42]{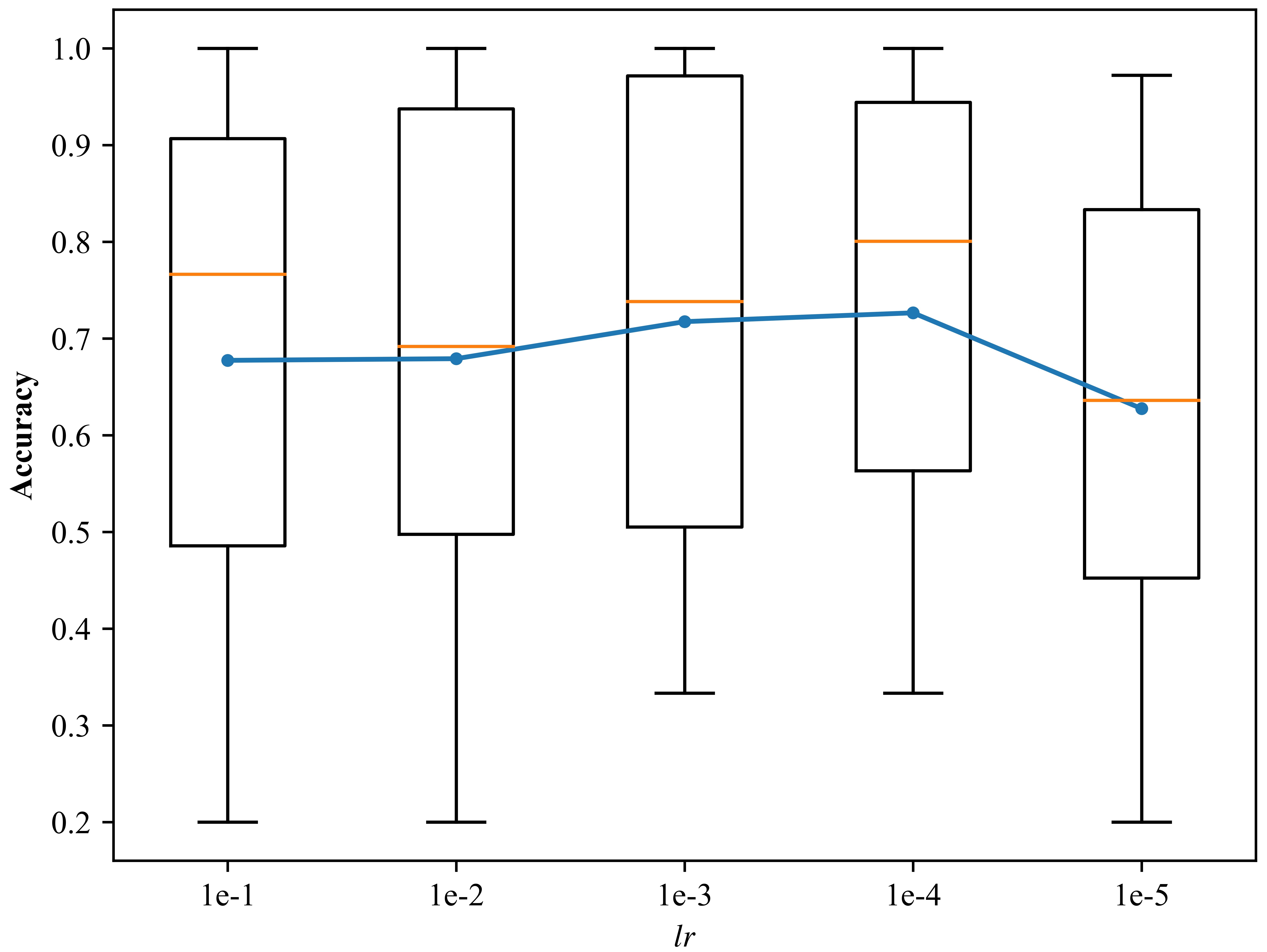}   
		\end{minipage}}
		\caption{Boxplot showing the accuracies on 12 UEA datasets vs. changes in the number of graphs $n_G$, pool ratio $\eta$, and learning rate.} 
		\label{fig:hyper}  
	\end{figure*}
	
	\subsection{Hyperparameter Stability}
	We conducted experiments on the 12 randomly selected datasets of the UEA benchmark by changing the number of dynamic graphs $n_G$, pooled ratio $\eta$, and learning rate $lr$ to evaluate the hyperparameter stability of TodyNet. Noteworthy, the number of dynamic graphs is corresponding to the number of slicing time series. \emph{AWR}, \emph{AF}, \emph{BM}, \emph{CR}, \emph{ER}, \emph{FM}, \emph{HMD}, \emph{HB}, \emph{NATO}, \emph{SRS1}, \emph{SRS2}, and \emph{SWJ} are chosen as the datasets for the experiments in this section. In Figure~\ref{fig:hyper}, the standard boxplots show the average classification accuracies of all datasets with respect to changes in $n_G:\{2,4,$ $6,8,10,12,14,16\}$, $\eta:$ $\{0.5,0.1,0.15,0.2,0.25,0.3,0.5\}$ and $lr:\{10^{-1},10^{-2},10^{-3},10^{-4}, 10^{-5}\}$. Note that the blue curve in the figure is the average accuracy of TodyNet over the selected 12 datasets with respect to the variation of parameters.
	
	Overall, the performance of the model remains at a high level. Apparently, the average accuracy tends to fluctuate slightly with the increase of $n_G$ due to a coarse-grained approach that these datasets with large length differences are split into the same number of time slots. However, in terms of overall performance, the implicit dependencies extracted by TodyNet are very effective for the time series classification task. It is obvious that the performance of the model first grows and then decreases to varying degrees with the changes of $\eta$, which is because the pooling structure will be flat if we reduce too many or very few nodes at one layer of our pooling approach. In addition, the results show that a suitable value of $lr$ is beneficial to obtain higher classification accuracy. Thus, we set appropriate parameters for different datasets to achieve the best performance of the model.
	
	\begin{table*}[t]
		\caption{Comparison of classification accuracy on 26 UEA benchmark.\label{tab:table2}}
		\centering
		\begin{tabular}{c c c c c c c c c c c}
			\hline
			\makecell{Datasets/Methods} & ED-1NN & \makecell{DTW\\-1NN-I} & \makecell{DTW\\-1NN-D} & \makecell{MLSTM\\-FCN} & ShapeNet & \makecell{WEASEL\\+MUSE} & TapNet & OS-CNN & MOS-CNN & TodyNet \\ \hline
			ArticularyWordRecognition & 0.970 & 0.980 & 0.987 & 0.973 & 0.987 & \underline{0.990} & 0.987 & 0.988 & \textbf{0.991} & 0.987 \\
			AtrialFibrillation & 0.267 & 0.267 & 0.200 & 0.267 & \underline{0.400} & 0.333 & 0.333 & 0.233 & 0.183 & \textbf{0.467} \\
			BasicMotions & 0.675 & \textbf{1.000} & \underline{0.975} & 0.95 & \textbf{1.000} & \textbf{1.000} & \textbf{1.000} & \textbf{1.000} & \textbf{1.000} & \textbf{1.000} \\
			Cricket & 0.944 & 0.986 & \textbf{1.000} & 0.917 & 0.986 & \textbf{1.000} & 0.958 & \underline{0.993} & 0.990 & \textbf{1.000} \\
			DuckDuckGeese & 0.275 & 0.550 & 0.600 & \underline{0.675} & \textbf{0.725} & 0.575 & 0.575 & 0.540 & 0.615 & 0.580 \\
			EigenWorms & 0.550 & 0.603 & 0.618 & 0.504 & \underline{0.878} & \textbf{0.890} & 0.489 & 0.414 & 0.508 & 0.840 \\
			Epilepsy & 0.667 & 0.978 & 0.964 & 0.761 & 0.987 & \textbf{1.000} & 0.971 & 0.980 & \underline{0.996} & 0.971 \\
			EthanolConcentration & 0.293 & 0.304 & 0.323 & \underline{0.373} & 0.312 & 0.133 & 0.323 & 0.240 & \textbf{0.415} & 0.350 \\
			ERing & 0.133 & 0.133 & 0.133 & 0.133 & 0.133 & 0.430 & 0.133 & \underline{0.881} & \textbf{0.915} & \textbf{0.915} \\
			FaceDetection & 0.519 & 0.513 & 0.529 & 0.545 & \underline{0.602} & 0.545 & 0.556 & 0.575 & 0.597 & \textbf{0.627} \\
			FingerMovements & 0.550 & 0.520 & 0.530 & \underline{0.580} & \textbf{0.589} & 0.490 & 0.530 & 0.568 & 0.568 & 0.570 \\
			HandMovementDirection & 0.279 & 0.306 & 0.231 & 0.365 & 0.338 & 0.365 & 0.378 & \underline{0.443} & 0.361 & \textbf{0.649} \\
			Handwriting & 0.371 & 0.509 & 0.607 & 0.286 & 0.451 & 0.605 & 0.357 & \underline{0.668} & \textbf{0.677} & 0.436 \\
			Heartbeat & 0.620 & 0.659 & 0.717 & 0.663 & \textbf{0.756} & 0.727 & \underline{0.751} & 0.489 & 0.604 & \textbf{0.756} \\
			Libras & 0.833 & 0.894 & 0.872 & 0.856 & 0.856 & 0.878 & 0.850 & \underline{0.950} & \textbf{0.965} & 0.850 \\
			LSST & 0.456 & 0.575 & 0.551 & 0.373 & \underline{0.590} & \underline{0.590} & 0.568 & 0.413 & 0.521 & \textbf{0.615} \\
			MotorImagery & 0.510 & 0.390 & 0.500 & 0.51 & \underline{0.610} & 0.500 & 0.590 & 0.535 & 0.515 & \textbf{0.640} \\
			NATOPS & 0.860 & 0.850 & 0.883 & 0.889 & 0.883 & 0.870 & 0.939 & \underline{0.968} & 0.951 & \textbf{0.972} \\
			PenDigits & 0.973 & 0.939 & 0.977 & 0.978 & 0.977 & 0.948 & 0.980 & \underline{0.985} & 0.983 & \textbf{0.987} \\
			PEMS-SF & 0.705 & 0.734 & 0.711 & 0.699 & 0.751 & N/A & 0.751 & 0.760 & \underline{0.764} & \textbf{0.780} \\
			PhonemeSpectra & 0.104 & 0.151 & 0.151 & 0.11 & 0.298 & 0.190 & 0.175 & \underline{0.299} & 0.295 & \textbf{0.309} \\
			RacketSports & 0.868 & 0.842 & 0.803 & 0.803 & 0.882 & \textbf{0.934} & 0.868 & 0.877 & \underline{0.929} & 0.803 \\
			SelfRegulationSCP1 & 0.771 & 0.765 & 0.775 & \underline{0.874} & 0.782 & 0.710 & 0.652 & 0.835 & 0.829 & \textbf{0.898} \\
			SelfRegulationSCP2 & 0.483 & 0.533 & 0.539 & 0.472 & \textbf{0.578} & 0.460 & \underline{0.550} & 0.532 & 0.510 & \underline{0.550} \\
			StandWalkJump & 0.200 & 0.333 & 0.200 & 0.067 & \textbf{0.533} & 0.333 & 0.400 & 0.383 & 0.383 & \underline{0.467} \\
			UWaveGestureLibrary & 0.881 & 0.869 & 0.903 & 0.891 & 0.906 & 0.916 & 0.894 & \textbf{0.927} & \underline{0.926} & 0.850 \\ \hline
			Average accuracy & 0.568 & 0.622 & 0.626 & 0.597 & 0.684 & 0.656 & 0.637 & 0.672 & \underline{0.692} & \textbf{0.726} \\
			Total best accuracy & 0 & 1 & 1 & 0 & 6 & 5 & 1 & 2 & 6 & 14 \\
			Ours 1-to-1-Wins & 24 & 20 & 19 & 19 & 13 & 16 & 19 & 19 & 16 & - \\
			Ours 1-to-1-Draws & 0 & 1 & 2 & 1 & 3 & 2 & 5 & 1 & 2 & - \\
			Ours 1-to-1-Losses & 2 & 5 & 5 & 6 & 10 & 8 & 2 & 6 & 8 & - \\ \hline
		\end{tabular}
	\end{table*}
	
	\subsection{Classification Performance Evaluation}
	To validate the performance of TodyNet, we compared the proposed TodyNet with baselines on all benchmark datasets in Table~\ref{tab:table2}, and the major results are shown in Table~\ref{tab:table2}. The specific accuracies of the baselines all refer to the original papers or~\cite{c:2}. The result marked with ``N/A" means that the corresponding method is unable to obtain the result due to memory or computational restriction. The best and second-best results are highlighted in bold and underlined, respectively.
	
	Table~\ref{tab:table2} indicates that TodyNet achieves the highest classification accuracy on 13 datasets. In terms of average accuracy, TodyNet shows the best performance and stability of 0.726 compared with the state-of-the-art baselines. For the comparisons of the 1-to-1-Wins/Draws/Losses, TodyNet performs much better than all the baselines. In particular, compared to state-of-the-art methods, the dynamic graph mechanism of TodyNet has a significant improvement for some datasets, such as \emph{FD}, \emph{FM}, \emph{HMD}, etc. Excitingly, these datasets belong to the type of ``EEG'' which is recorded from magnetoencephalography (MEG) and contains the tags of human behavior. It is easy to see that the hidden dependencies between brain signals at different locations jointly determine human behavior, and TodyNet is able to capture such spatio-temporal features well and shows great potential. Simultaneously, it indicates that the dynamic characteristics of time series also positively influence the results of the classification.
	
	\begin{figure}[!h]
		\centering
		\includegraphics[width=1\columnwidth]{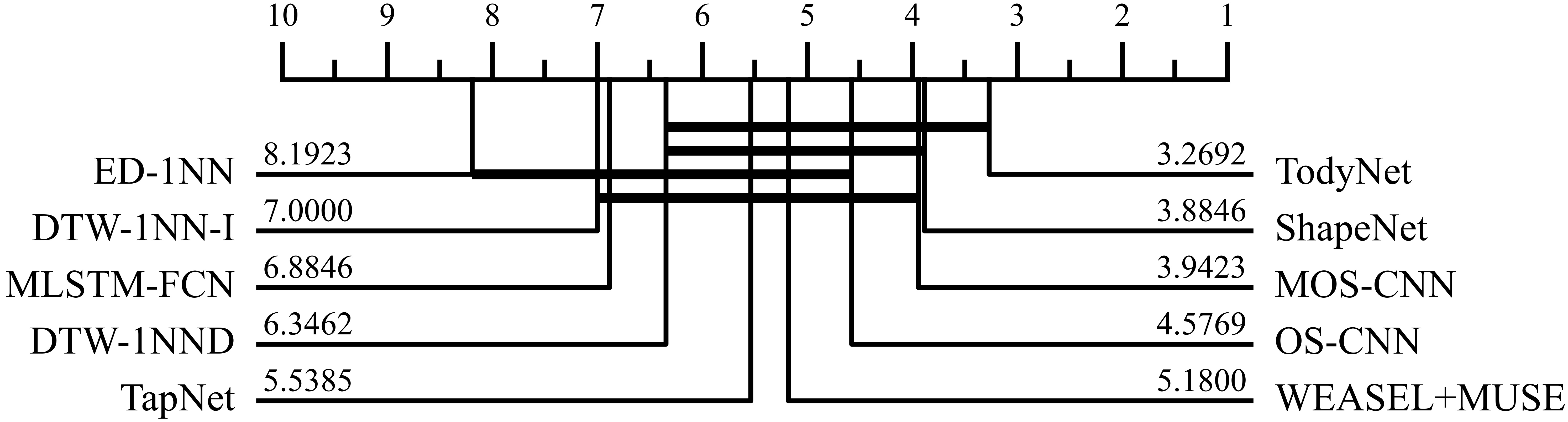}
		\caption{Critical difference diagram on the 26 UEA datasets with 
			$\alpha = 0.05$.}
		\label{fig:cd}
	\end{figure}
	
	On the other hand, we also conducted the Wilcoxon signed-rank test to evaluate the performances of all approaches. Figure~\ref{fig:cd} shows the critical difference diagram with $\alpha = 0.05$ which is plotted according to the results in Table \ref{tab:table2}. The values in Figure~\ref{fig:cd} also reflect the average performance rank of the classifiers. TodyNet achieves the best overall average rank of 3.2692, which is lower than the ranks of the existing state-of-the-art deep learning-based approaches, such as MOS-CNN, WEASEL+MUSE, and TapNet. Besides, TodyNet is significantly better than other classifiers, e.g., ShapeNet, MLSTM-FCN, and DTW. This is because the extraction of hidden dynamic dependencies can significantly improve the performance of classification.
	
	\begin{figure*}[!b]
		\centering    
		{\begin{minipage}{0.32\textwidth}
				\centering    
				\includegraphics[scale=0.42]{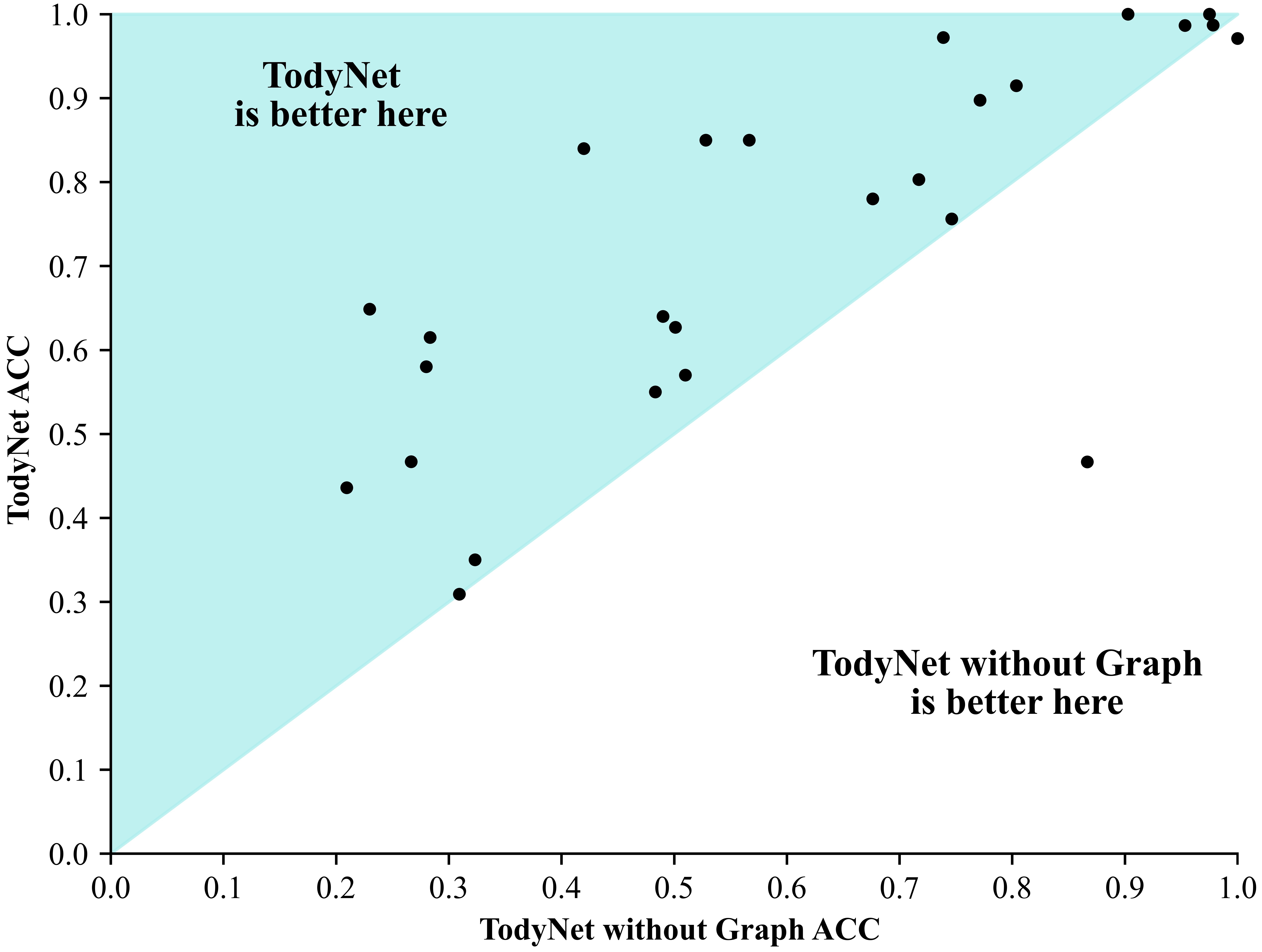}   
		\end{minipage}}	
		{\begin{minipage}{0.32\textwidth}
				\centering    
				\includegraphics[scale=0.42]{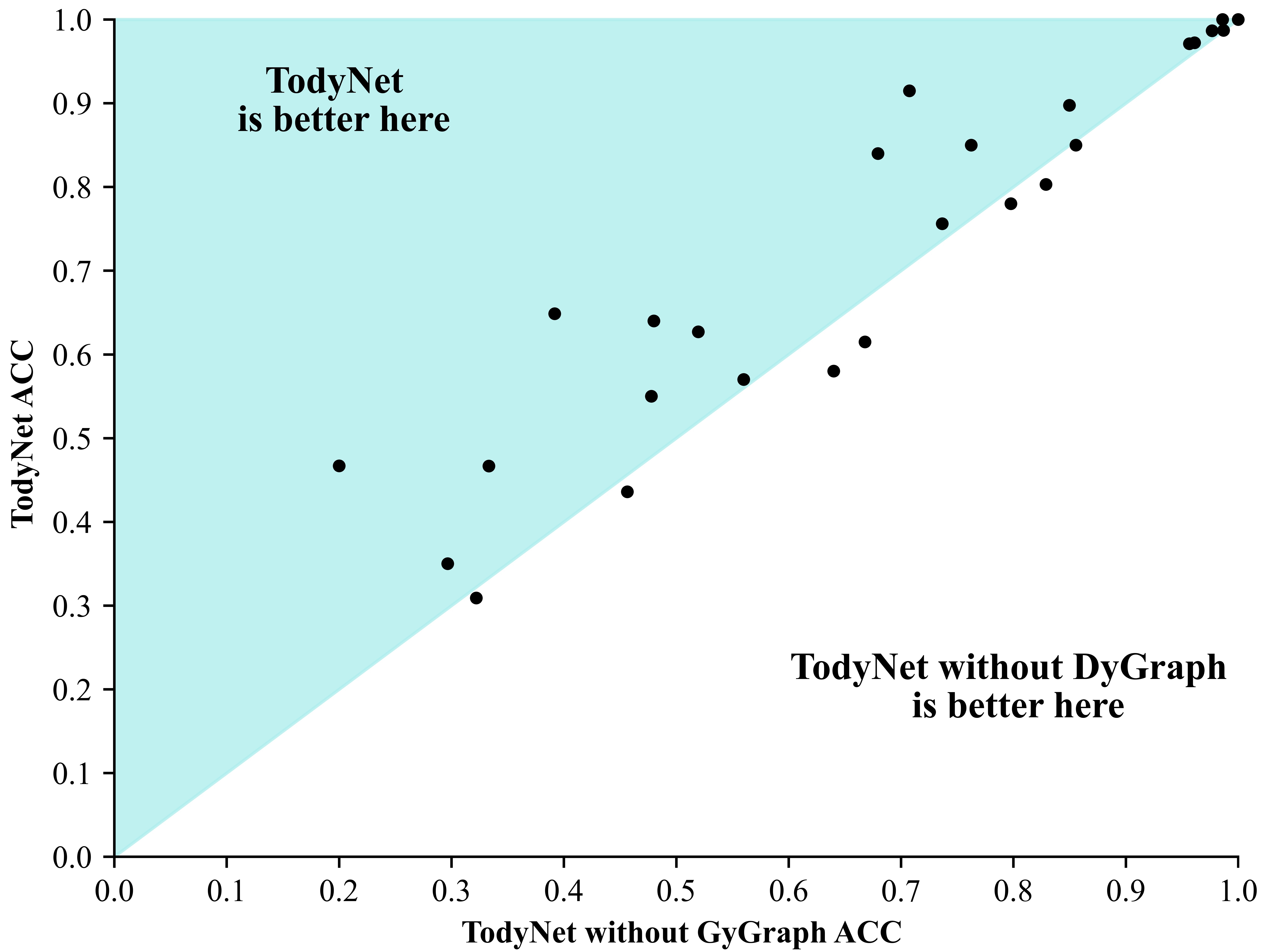}   
		\end{minipage}}	
		{\begin{minipage}{0.32\textwidth}
				\centering    
				\includegraphics[scale=0.42]{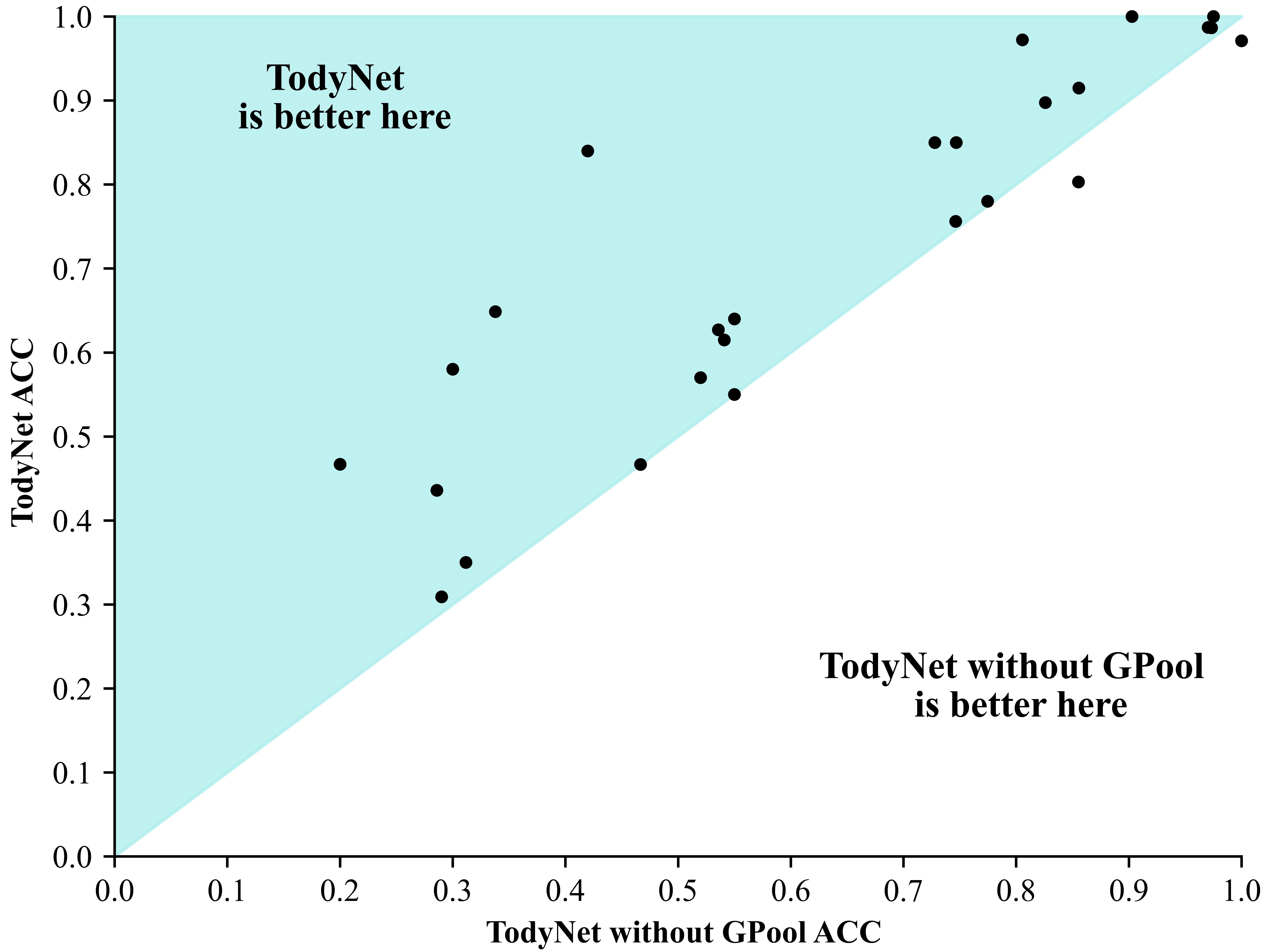}   
		\end{minipage}}	
		
		\caption{Scatter plot for TodyNet, TodyNet without graph mechanism, dynamic graphs, and TodyNet without temporal graph pooling on the 26 UEA MTSC problems.} 
		\label{fig:ablation}  
	\end{figure*}
	
	\subsection{Ablation Study}
	In order to validate the effectiveness of the key components that contribute to the improved
	outcomes of TodyNet, we employ ablation studies on the 26 UEA multivariate time series archive. We denote TodyNet without different components as follows:
	
	\begin{itemize}
		\item \textbf{w/o Graph}: TodyNet without temporal graph and graph neural networks. We pass the outputs of the temporal convolution to the output layer directly.
		\item \textbf{w/o DyGraph}: TodyNet without the temporal dynamic graph. We only construct one static graph as one of the inputs of graph neural networks by using the mentioned graph construction method and do not slice the time series.
		\item \textbf{w/o GPool}: TodyNet without the temporal graph pooling. We directly concatenate the outputs of the graph neural networks and pass them to the output module.
	\end{itemize}
	
	The scatter plots are shown in Figure~\ref{fig:ablation} that demonstrate the experimental comparison results of reducing different components of TodyNet. The horizontal and vertical coordinates of Figure~\ref{fig:ablation} indicate the classification accuracy, and the points falling in the pale turquoise region indicate that ToyNet is more accurate on the corresponding dataset.
	
	From Figure~\ref{fig:ablation}, a conclusion can be drawn from the results that the graph information significantly improves the outcomes of convolution-based classifier relying on the excellent capability of TodyNet to capture the hidden dependencies among variables and dynamic associations between different time slots. We can also observe that the introduction of a dynamic graph mechanism can obtain better model performance than a static graph because it enables information to flow between isolated but interdependent time slots. In addition, the effect of temporal graph pooling is evident as well: it validates that temporal graph pooling helps aggregate hub data in a hierarchical way which enhances the performance of graph neural networks to a great extent. In a word, all results of ablation experiments manifest that the proposed components of TodyNet are all indeed effective for the multivariate time series classification tasks.
	
	\subsection{Inspection of Class Prototype}
\begin{figure}[!h]
		\centering
		\includegraphics[width=1\columnwidth]{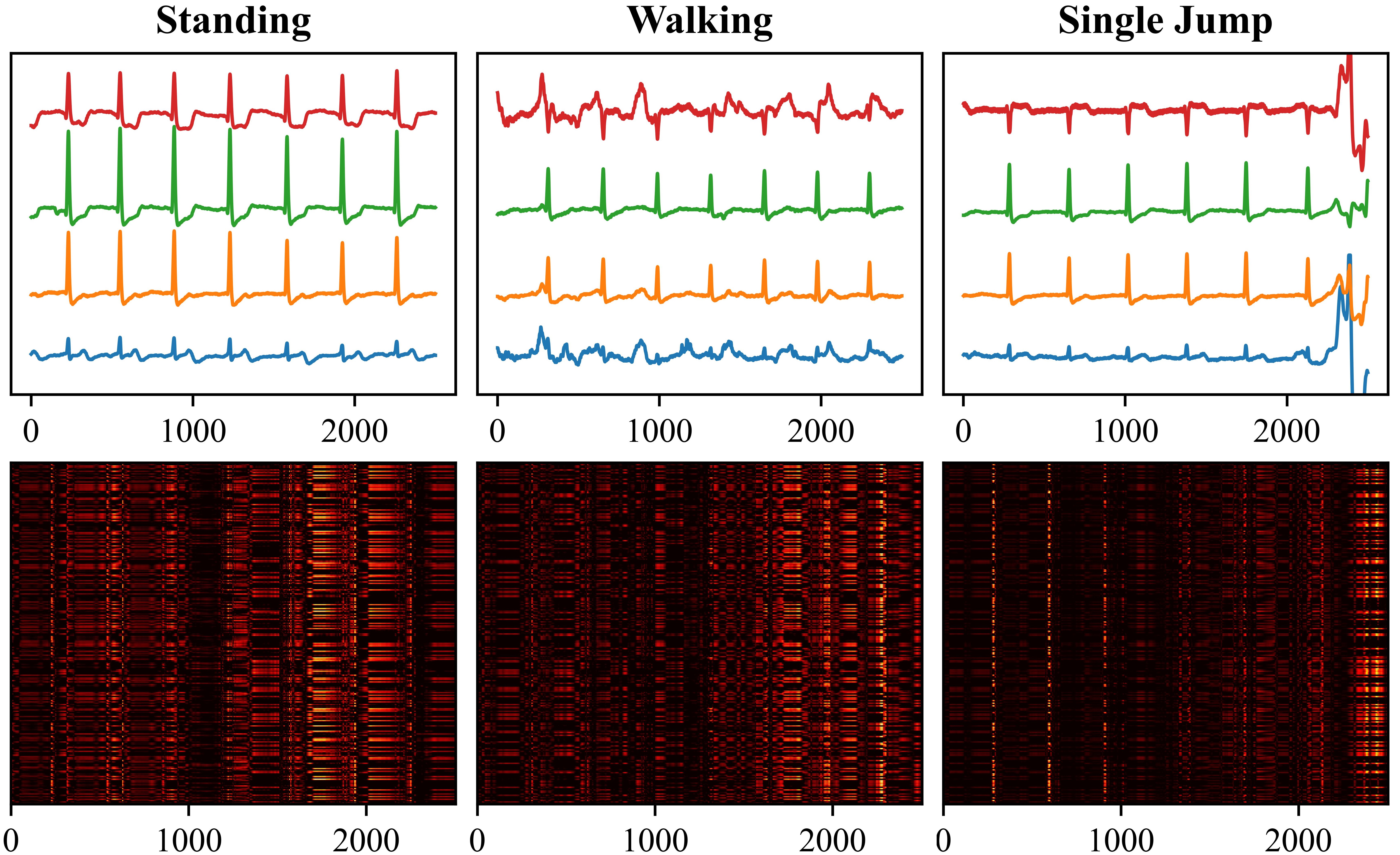}
		\caption{Heatmap visualization of representations learned by TodyNet on the StandWalkJump} dataset.
		\label{fig:hm}
	\end{figure}

	In this experiment, we show the effectiveness of our well-trained time series embedding by visualizing the class prototype and its corresponding time series embedding. To begin with, we analyze the representation learned by TodyNet over epochs with a heatmap. The \emph{SWJ} is a dataset of UEA archive that records short-duration ECG signals of 4 pairs of electrodes from a healthy male performing 3 different physical activities: Standing, Walking, and Single Jump. We randomly selected 3 samples which are individually corresponding to one category from 15 test samples of \emph{SWJ} for heatmap visualization. In Figure~\ref{fig:hm}, each row of subplots shows the original signal of all sensors and the learned embedding from each sample under the corresponding class, respectively. The outcomes indicate that there are significant differences in the embeddings learned from different categories. Thus, an obvious conclusion is that TodyNet can clearly distinguish the classes of different samples of time series by learning efficient representations.
	
\begin{figure}[!h]
		\centering
		\includegraphics[width=1\columnwidth]{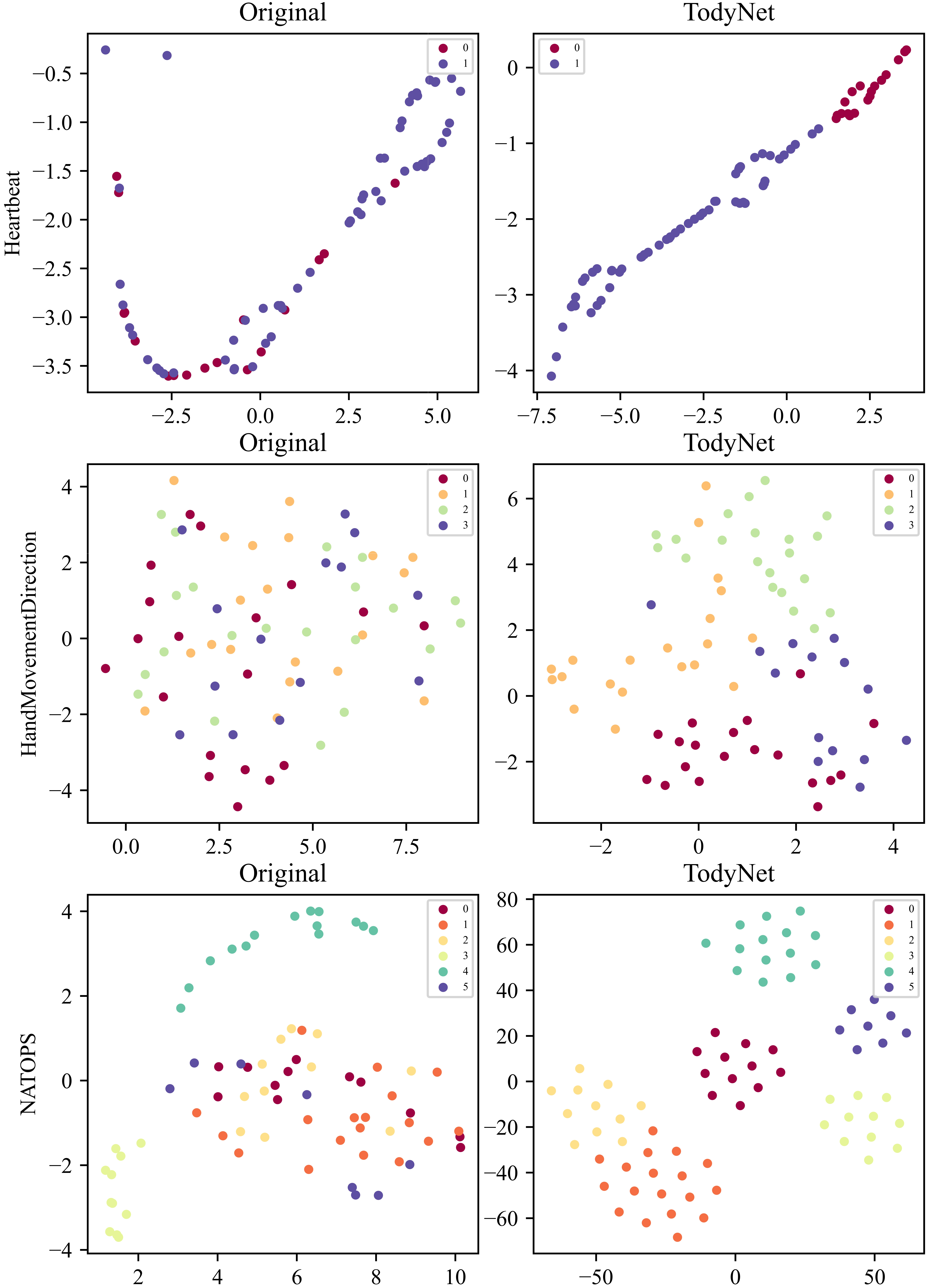}
		\caption{The t-SNE visualization of the representation space for the datasets Heartbeat, HandMovementDirection, and NATOPS.}
		\label{fig:icp}
	\end{figure}

	To further demonstrate the ability of TodyNet to represent time series, we use the t-SNE algorithm~\cite{van2008visualizing} to embed the representation into the form of a two-dimensional image for visualization. Figure~\ref{fig:icp} shows the results on test datasets of \emph{HB}, \emph{HMD}, and \emph{NATO}. For convenience, all class labels are unified by Arabic numerals. As shown in Figure~\ref{fig:icp}, each row denotes different datasets, and each column represents the visualization of the original data and the embeddings learned by TodyNet, respectively. It is clear that the distance between different samples from the same class is closer, which implies that TodyNet can effectively characterize class prototypes and enables highly accurate classification for different data.
	
	\section{Conclusion}
	In this paper, a novel framework named temporal dynamic graph neural network was proposed. To the best of our knowledge, this is the first dynamic graph-based deep learning method to address multivariate time series classification problems. We propose an effective method to extract the hidden dependencies among multiple time series and the temporal dynamic feature among different time slots. Meanwhile, a novel temporal graph pooling method is designed to overcome the flat of the graph neural network. Our method has shown superb performance compared with other state-of-the-art methods on UEA benchmarks. For future directions, there are two possible ways. On the one hand, a better classification performance may be achieved by transplanting the other temporal convolution methods. On the other hand, reducing the complexity of temporal graph pooling can further improve the efficiency of the model.
	
	
	\bibliographystyle{IEEEtran}
	\bibliography{mybibliography}
	
	\newpage
	
	
	
	
	

	\vfill
	
\end{document}